\documentclass[letterpaper, 10 pt, conference]{ieeeconf}

\IEEEoverridecommandlockouts
\usepackage{amsmath}
\usepackage{amssymb}
\usepackage{bm}
\usepackage{booktabs}
\usepackage{graphicx}
\usepackage{url}
\usepackage{algorithm}
\usepackage{algorithmic}
\usepackage{balance}
\usepackage{hyperref}

\newtheorem{remark}{Remark}

\title{\LARGE \bf
Neural Backward Reach-Avoid Tubes with MPC Supervision for High-Dimensional Systems: An Application to Safe Spacecraft Docking
}

\author{Santiago Thorup$^{1,*}$, Luca Castelletto$^{1,*}$, Zeyuan Feng$^{1}$, and Somil Bansal$^{1}$%
\thanks{$^{*}$Equal contribution}%
\thanks{$^{1}$Department of Aeronautics and Astronautics at Stanford University, USA: \{sthorup4, lucacast, zeyuanf, somil\}@stanford.edu}%
\thanks{The implementation can be found on \texttt{https://github.com/santiagothorup/reachAvoidDocking}}
}

\begin{document}

\maketitle
\thispagestyle{empty}
\pagestyle{empty}

\begin{abstract}
Autonomous spacecraft docking requires control policies that simultaneously ensure collision avoidance and target reachability under coupled, high-dimensional translational–rotational dynamics. Hamilton–Jacobi (HJ) reachability provides formal reach–avoid guarantees, but classical solvers are limited to low-dimensional systems. Learning-based approaches have begun to scale HJ analysis, yet they struggle in reach-avoid settings, especially where goal and failure sets are tightly coupled, as in docking.
We propose a learning-based Backward Reach-Avoid Tube (BRAT) framework that addresses this challenge by tightly integrating HJ structure with MPC-based supervision. In the offline phase, we train a neural approximation of the HJ value function using PDE-based losses augmented with curriculum-driven MPC supervision, which provides informative value targets and stabilizes training in regions where purely PDE-based methods fail. In the online phase, the learned value function is deployed through two real-time controllers: (i) a value gradient-driven controller, and (ii) a value-function-augmented terminal MPC that explicitly enforces reachability at the horizon.
We evaluate the proposed method on a 6D planar docking problem against grid-based ground truth and then scale to the full 13D system. Across both settings, our approach outperforms existing methods in success rate and computational efficiency. 
\end{abstract}

\begin{figure*}[!ht]
  \centering
  \includegraphics[width=0.95\linewidth, trim=0 0cm 0 0, clip]{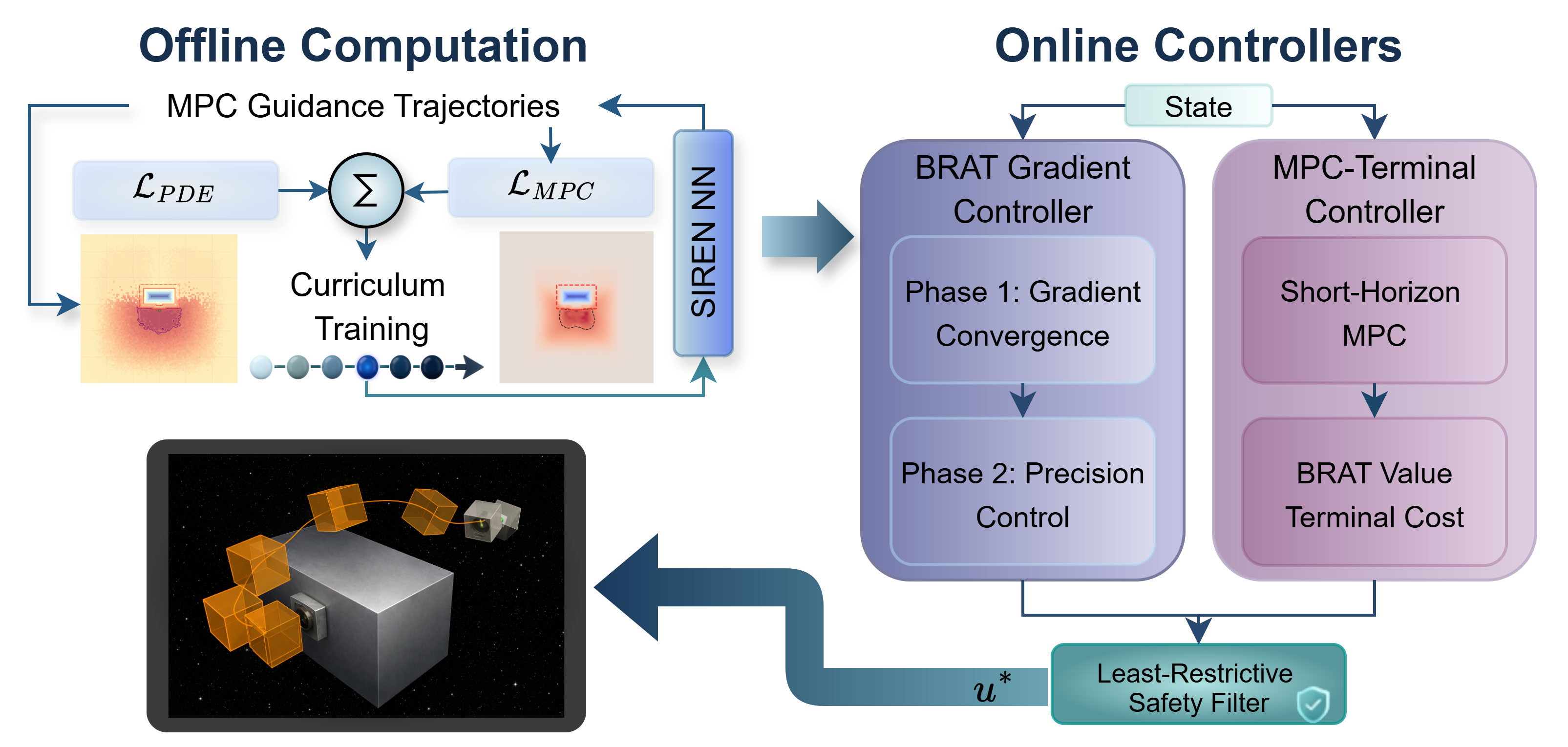}
  \caption{Overview of the learning-based BRAT framework. \textbf{Top-Left (Offline):} MPC-generated guidance trajectories provide supervision labels that, combined with the PDE residual loss, train a neural network through a curriculum learning. \textbf{Right (Online):} An online BRAT gradient controller operating in convergence and precision phases, and an online MPC-terminal controller using the value function as a terminal cost are both backed by a least-restrictive safety filter to ensure safety. \textbf{Bottom-left:} Example 13D docking trajectories demonstrating a safe approach to the target.}
  \label{fig:overall_sum}
\end{figure*}

\section{Introduction}
Autonomous spacecraft docking is a critical capability for on-orbit servicing, debris removal, and space station resupply~\cite{fehse2003}. As missions grow in complexity,
guaranteed safety during proximity operations has become a fundamental requirement~\cite{Petersen2023}. A successful docking controller must steer the chaser into a small capture window while avoiding a collision with the target body, under tightly coupled translational and rotational dynamics.

Classical approaches rely on optimization-based control, including LQR~\cite{jewison2016}, fuel-optimal convex programming~\cite{malyuta2022}, and MPC-based trajectory optimization~\cite{eren2017}. While effective, these methods incur significant online computational cost, especially for power and compute-gated satellites, and lack global safety guarantees. Reinforcement learning approaches~\cite{hovell2021, broida2019} improve execution speed but similarly do not provide formal safety certificates. Safety filter architectures~\cite{wabersich2021, ames2019control} can enforce safety constraints on a nominal controller, but do not guarantee task completion and can lead to degraded performance.

Hamilton–Jacobi (HJ) reachability provides a principled framework for certifying \textit{reach–avoid} behaviors~\cite{mitchell2005, bansal2017hamilton}. The resulting value function characterizes all states from which the system can safely reach the goal -- also referred to as the \textit{backward
reach-avoid tube (BRAT)} of the system. 
The value function gradient yields the optimal feedback policy. While HJ methods have been applied to spacecraft proximity operations in low dimensions~\cite{chaudhuri2016, zagaris2018}, classical grid-based solvers suffer from the curse of dimensionality, limiting their practical use to five or six dimensions~\cite{mitchell2005, lygeros2004}.
This limitation is particularly restrictive for spacecraft docking, where full translational–rotational dynamics lead to a 13D system.

Learning-based methods aim to overcome this limitation by approximating the HJ value function with neural networks~\cite{bansal2021deepreach, hsu2021safety}. For instance, DeepReach~\cite{bansal2021deepreach} leverages sinusoidal representations~\cite{sitzmann2020siren} to minimize HJ PDE residuals, and subsequent work has shown that augmenting this objective with MPC-generated value targets can significantly improve training robustness~\cite{MPCDeepReach}. However, these approaches have primarily been demonstrated on avoid-only problems and struggle in reach–avoid settings with tightly coupled goal and failure sets.
Practically speaking, reach–avoid problems introduce additional learning challenges beyond avoid-only settings. The associated HJ PDE (or variational inequality) involves nested min–max operators that produce flat gradients away from the active boundary, leaving large regions of the state space weakly supervised. Moreover, BRAT typically expands over time, requiring the learned value function to accurately track a moving and growing decision boundary -- unlike backward reachable tubes (BRTs), which typically converge within short horizons. These difficulties make naïve PDE-based training unstable and often lead to suboptimal or unsafe policies.

In this work, we address these challenges by tightly integrating MPC-based supervision into the learning of reach–avoid value functions. Our framework  combines PDE residual-based training with curriculum-driven MPC-generated value labels and a train–verify–refine procedure to enable stable learning in high-dimensional systems. The learned value function is then deployed through a suite of real-time controllers for safe and efficient docking.
To summarize, our contributions are as follows: 
\begin{enumerate}
\item We extend MPC-supervised DeepReach~\cite{MPCDeepReach} to reach–avoid problems, introducing a curriculum-based training strategy with a \emph{train-verify–refine} loop for stable long-horizon value function learning.
\item We develop two real-time controllers from the learned value function for deployment: a BRAT-gradient-driven controller and a value-function-augmented terminal MPC, both supported by a least-restrictive safety filter to ensure safety.
\item We evaluate our approach on spacecraft docking across models of increasing fidelity: (a) a 6D system with grid-based ground truth comparison, and (b) the full 13D system, with goal sets derived from IDSS tolerances~\cite{nasaidss}.
\item We show that our method outperforms MPC, DeepReach~\cite{bansal2021deepreach}, and RL baselines~\cite{hsu2021safety} in both success rate and computational efficiency, while achieving faster docking than grid-based solutions.
\end{enumerate}

\section{Problem Statement}
Consider a controlled dynamical system $\dot{x} = f(x, u)$ with state $x \in \mathbb{R}^n$, control $u \in \mathcal{U}$, a goal set $\mathcal{G} = \{x : g(x) \leq 0\}$ defined by a Lipschitz-continuous reach function $g$, and a failure set $\mathcal{L} = \{x : \ell(x) \leq 0\}$ defined by an avoid function $\ell$. Given a problem horizon $T > 0$, our objectives are twofold.

\textit{1) Compute the Backward Reach-Avoid Tube (BRAT):} the set 
of all initial states from which the system can reach $\mathcal{G}$ 
while avoiding $\mathcal{L}$ within horizon $T$:
\begin{multline}\label{eq:brat}
\mathcal{R}(T) = \bigl\{ x_0 \mid
\exists\, u(\cdot) \in \mathcal{U}^{[0,T]},\;
\exists\, s \in [0,T]: \\
\xi_{x_0}^u(s) \in \mathcal{G},\;\;
\xi_{x_0}^u(\tau) \notin \mathcal{L}\;\;
\forall\, \tau \in [0, s] \bigr\},
\end{multline}
where $\xi_{x_0}^u(\cdot)$ is the trajectory from $x_0$ under 
control $u(\cdot)$.

\textit{2) Synthesize the optimal reach-avoid policy:} 
Although $\mathcal{R}(T)$ is necessarily computed over a finite horizon $T$, we seek a policy $u^*(x)$ that approximates the infinite-horizon reach-avoid behavior, driving the system along a collision-free trajectory to the goal from initial states $x_0 \in \mathcal{R}(\infty)$ even when they lie outside the $T$-horizon tube.


In this work, we apply these objectives to autonomous spacecraft docking, where a chaser satellite must navigate into a narrow capture window while avoiding collision with a target satellite in a 400\,km circular orbit. The goal set encodes simultaneous position, velocity, attitude, and angular rate tolerances for a successful dock, while the failure set is the target body inflated by a collision buffer. This problem is particularly challenging for learning-based reachability because the goal and failure sets are tightly coupled in physical space---the chaser must pass close to the obstacle to reach the docking port---and the coupled translational-rotational dynamics span 13 dimensions, far beyond the reach of grid-based solvers, necessitating the development of more scalable reachability methods for reach-avoid problems. 
\section{Preliminaries}

\subsection{HJ Value Function for BRATs}

Given a trajectory $\xi_{x_0}^u(\cdot)$ from initial state $x_0$ at time $t$ under control $u(\cdot)$, the reach-avoid cost functional encodes whether the system reaches the goal without entering the failure set:
\begin{equation}\label{eq:cost}
J(x, u(\cdot), t) = \min_{\tau_1 \in [t,T]} \max\!\Big\{g\!\big(\xi(\tau_1)\big),\; \max_{\tau_2 \in [t,\tau_1]} \big\{-\ell\!\big(\xi(\tau_2)\big)\big\}\Big\}.
\end{equation}
The outer minimization over $\tau_1$ searches for the best time to reach the goal, while the inner maximization over $\tau_2$ penalizes any failure-set violation prior to arrival. 

The value function is the optimal cost achievable over all admissible controls:
\begin{equation}\label{eq:valfun}
V(x, t) = \min_{u(\cdot) \in \mathcal{U}^{[t,T]}} J(x, t, u(\cdot)).
\end{equation} 

The value function satisfies the reach-avoid variational inequality (VI)~\cite{fisac2015, margellos2011}:
\begin{equation}\label{eq:hjvi}
\min\!\Big\{\max\!\Big\{\frac{\partial V}{\partial t} + H(x, \nabla_x V),\; V - g(x)\Big\},\; V + \ell(x)\Big\} = 0,
\end{equation}
for $t \in [0, T)$, with the boundary condition $V(x, T) = \max\!\big(g(x),\; -\ell(x)\big)$, which is non-positive if and only if $x \in \mathcal{G}$ and $x \notin \mathcal{L}$. The Hamiltonian is given as
\begin{equation}\label{eq:hamiltonian}
H(x, p) = \min_{u \in \mathcal{U}} \; p^\top f(x, u),
\end{equation}
and $p = \nabla_x V$ is the costate. The inner $\max$ in~\eqref{eq:hjvi} ensures either the PDE evolution holds or $V$ is capped by the reach function when the goal has already been reached. The outer $\min$ ensures $V \geq -\ell$, enforcing obstacle avoidance.

The zero sub-level set of $V$ recovers the BRAT: $\mathcal{R}(t) = \{x : V(x, t) \leq 0\}$, and its gradient yields the optimal control. The optimal reach-avoid policy selects the control that drives the system toward the goal as quickly as possible while remaining safe. For control-affine dynamics $f(x, u) = f_0(x) + B(x)u$, minimizing the Hamiltonian~\eqref{eq:hamiltonian} yields a closed-form bang-bang solution:
\begin{equation}\label{eq:bangbang}
u_i^* = -\bar{u}_i \cdot \mathrm{sign}\!\big((B(x)^\top \nabla_x V)_i\big).
\end{equation}

\subsection{DeepReach}\label{sec:deepreach}

DeepReach~\cite{bansal2021deepreach} approximates the HJ value function with a SIREN neural network~\cite{sitzmann2020siren} $\phi_\theta(x, t)$ trained by minimizing the PDE residual of~\eqref{eq:hjvi} at sampled collocation points. By replacing the exponentially scaling grid with a fixed-size network, DeepReach circumvents the curse of dimensionality and has been demonstrated on systems up to 10 dimensions. However, for reach-avoid problems, the nested min-max structure of the VI produces vanishingly small training gradients in regions far from the active boundary, causing slow and often inaccurate convergence---particularly when the goal and failure sets are geometrically complex or tightly coupled. These limitations motivate the MPC-supervised training and curriculum strategies developed in this work.

\section{Method}

\subsection{Approach Overview}
We build upon DeepReach~\cite{bansal2021deepreach} to train a neural value function that approximates the HJ reach-avoid solution~\eqref{eq:hjvi} in high dimensions. However, standard DeepReach training fails on the reach-avoid docking problem for two reasons. First, the nested min-max structure of the VI produces vanishingly small training gradients over most of the state space, leaving the network under-constrained far from the 
active boundary (\emph{spatial challenge}). Second, learning the full-horizon value function from the start causes errors at short horizons to compound into longer-horizon solutions, destabilizing training (\emph{temporal challenge}).

Our framework addresses these through two complementary innovations. To solve the spatial challenge, we augment the PDE loss with MPC-generated reach-avoid cost labels that provide 
direct supervision where the VI residual gives a weak signal, extending the MPC-supervised framework of~\cite{MPCDeepReach} from avoid-only to reach-avoid semantics. To solve the temporal challenge, we introduce a \emph{train-verify-refine} cycle that gates curriculum advancement on validation against MPC cost estimates and regenerates labels using the current policy as a warm start, creating a feedback loop between network quality and supervision quality. These two innovations are tightly coupled: MPC labels without verification accumulate stale supervision at longer horizons, while the verification cycle without MPC labels has no meaningful signal to validate against. To deploy the learned value function online, we develop two controllers tailored to the reach-avoid structure, discussed in \ref{sec:controllers}.


\begin{algorithm}[!htbp]
\caption{MPC-Supervised Neural BRAT Training}\label{alg:training}
\begin{algorithmic}[1]
\REQUIRE Dynamics $f$, target function $g(x)$, avoid function $l(x)$, time horizon $T$
\STATE Initialize SIREN network $\phi_\theta$
\STATE Generate initial MPC cost labels
\FOR{epoch $= 1, 2, \ldots, N_C$}
  \STATE $t_{\max} \leftarrow \mathrm{curriculum}(\text{epoch})$ \COMMENT{Eq.~(\ref{eq:curriculum})}
  \STATE Sample PDE collocation points and MPC-labeled points
  \STATE Step optimizer on $w_{\mathrm{pde}}\,\mathcal{L}_{\mathrm{pde}} + w_{\mathrm{mpc}}\,\mathcal{L}_{\mathrm{mpc}}$ \COMMENT{Eqs.~(\ref{eq:pde_loss}--\ref{eq:mpc_loss})}
  \IF{curriculum checkpoint reached}
    \STATE Validate $V_\theta$ on held-out $x$; advance if converged
    \STATE Regenerate MPC with policy as warm-start
\ENDIF
\ENDFOR
\end{algorithmic}
\end{algorithm}

\subsection{Offline Training}\label{sec:training}
We aim to learn the reach-avoid value function with a SIREN network~\cite{sitzmann2020siren} $\phi_\theta(x, t)$ and an exact-boundary-condition parameterization~\cite{singh2025exact, lagaris1998}:
\begin{equation}\label{eq:exact}
V_\theta(x, t) = \phi_\theta(x, t) \cdot (T - t) + \max\!\big(g(x),\; -\ell(x)\big).
\end{equation}
The factor $(T - t)$ guarantees boundary condition satisfaction (i.e., $V_\theta(x,T)=\max\!\big(g(x),\; -\ell(x)\big)$ for any network weights, eliminating the boundary condition loss ---particularly important when the goal set occupies a vanishingly small fraction of the state space. 

The network is trained by minimizing a combined loss:
\begin{equation}\label{eq:loss}
\mathcal{L} = w_{\mathrm{pde}} \cdot \mathcal{L}_{\mathrm{pde}} + w_{\mathrm{mpc}} \cdot \mathcal{L}_{\mathrm{mpc}}.
\end{equation}
The \emph{PDE loss} penalizes violations of the VI~\eqref{eq:hjvi}:
\begin{equation}\label{eq:pde_loss}
\mathcal{L}_{\mathrm{pde}} = \frac{1}{N_p}\sum_{i=1}^{N_p} \left| \min\!\left\{\max\!\left\{\frac{\partial V}{\partial t} + H,\; V - g\right\},\; V + \ell\right\}\right|.
\end{equation}
The \emph{MPC supervision loss} provides direct value-function labels generated via perturbation-based shooting~\cite{MPCDeepReach}:
\begin{equation}\label{eq:mpc_loss}
\mathcal{L}_{\mathrm{mpc}} = \frac{1}{N_m}\sum_{j=1}^{N_m} |V_\theta(x_j^m, t_j^m) - V_j^{\mathrm{mpc}}|,
\end{equation}
with an additional false-positive penalty when $V_j^{\mathrm{mpc}} > 0$ but $V_\theta \leq 0$, discouraging the network from labeling unsafe states as reachable. The MPC labels are generated using the sampling-based solver from \cite{MPCDeepReach}, adapted to utilize the reach-avoid cost function defined in \eqref{eq:cost}. Labels are concentrated near the goal via importance sampling across four strata, ranging from near-goal perturbations to broad uniform coverage. 

\subsection{Train-Verify-Refine Cycle}\label{sec:curriculum}

Standard DeepReach employs a linear curriculum that gradually expands the collocation time range from $[T, T]$ to $[0, T]$ during training. We retain this schedule:
\begin{equation}\label{eq:curriculum}
t_{\min}(\text{epoch}) = T - T \cdot \alpha \cdot \min\!\left(\frac{\text{epoch}}{N_c},\; 1\right),
\end{equation}
where $\alpha > 1$ provides a slight overshoot for stability and $N_c$ is the total number of training epochs. However, a passive linear schedule is insufficient for reach-avoid problems: the BRAT grows monotonically with the horizon, so errors at short horizons propagate into and corrupt longer-horizon solutions if the curriculum advances prematurely.

We replace this passive schedule with an active feedback loop. The horizon is divided into equally spaced checkpoints. At each checkpoint, training pauses and the current value function is evaluated against MPC cost estimates on a held-out set of states. If the validation error exceeds a threshold, training continues at the current horizon until accuracy improves. Only once validation passes does the curriculum advance to the next checkpoint.

Crucially, at each checkpoint, we also regenerate the MPC supervision labels. Rather than re-solving from scratch, the MPC trajectories are warm-started from the current learned policy, so that as the network improves, the quality of the supervision labels improves in tandem. This creates a bootstrapping loop: a better value function produces better warm-start trajectories, which yield more accurate MPC labels, which further refine the value function. Without this coupling, supervision labels generated early in training become increasingly stale as the horizon grows, degrading convergence. We find this feedback mechanism critical for stable training at the 13D scale.

\subsection{Online Controllers}\label{sec:controllers}

Because the training horizon $T$ is finite, states outside the learned BRAT ($V(x, 0) > 0$) have no formal guarantee of reaching the goal within $T$. However, the value function gradient still provides useful control information beyond the BRAT boundary. Our controllers exploit this by operating in two phases, augmented by a safety filter.

\subsubsection{BRAT Controller}
\emph{Phase~1 (Convergence):} When $V(x, 0) > 0$, the state lies outside the trained BRAT. The controller queries the value function at $t = 0$ (full horizon) and applies bang-bang control~\eqref{eq:bangbang} using the gradient direction, steering the chaser toward the BRAT interior. Although the value function does not certify reachability in this region, the gradient still points toward decreasing $V$, providing a meaningful descent direction. In regions where the value function is locally flat, PD feedback toward the goal state is blended in to maintain control authority.

\emph{Phase~2 (Precision):} Once $V(x, 0) \leq 0$, the state has entered the BRAT and safe docking is achievable within $T$. The controller performs a minimum-time search: it finds the largest $t^*$ such that $V(x, t^*) \leq 0$ and applies bang-bang control at this tighter time slice. As the chaser approaches docking, $t^*$ increases toward $T$, progressively narrowing the effective horizon and yielding more precise terminal guidance. The transition from Phase~1 to Phase~2 is one-way.

\subsubsection{Terminal MPC Controller}
This controller uses short-horizon MPC with the learned value function $V(x, t)$ as a differentiable terminal cost. The terminal cost transitions from $V(x, 0)$ in Phase~1 to $V(x, t^*)$ in Phase~2 via the same minimum-time search, combining the computational tractability of short-horizon planning with the long-horizon reach-avoid information encoded in $V$.

\emph{Least-Restrictive Safety Filter}
In Phase~1, the chaser operates outside the BRAT where the controller does not guarantee collision-free behavior. A separately trained avoid-only BRT provides a least-restrictive safety layer: when $V_{\mathrm{avoid}}(x)$ drops below a margin $\gamma$, the nominal control is overridden with bang-bang that maximizes $V_{\mathrm{avoid}}$, driving the chaser away from the obstacle. The filter intervenes only when necessary, preserving the nominal controller's authority otherwise.


\begin{figure}
    \centering
    \includegraphics[width=1\linewidth]{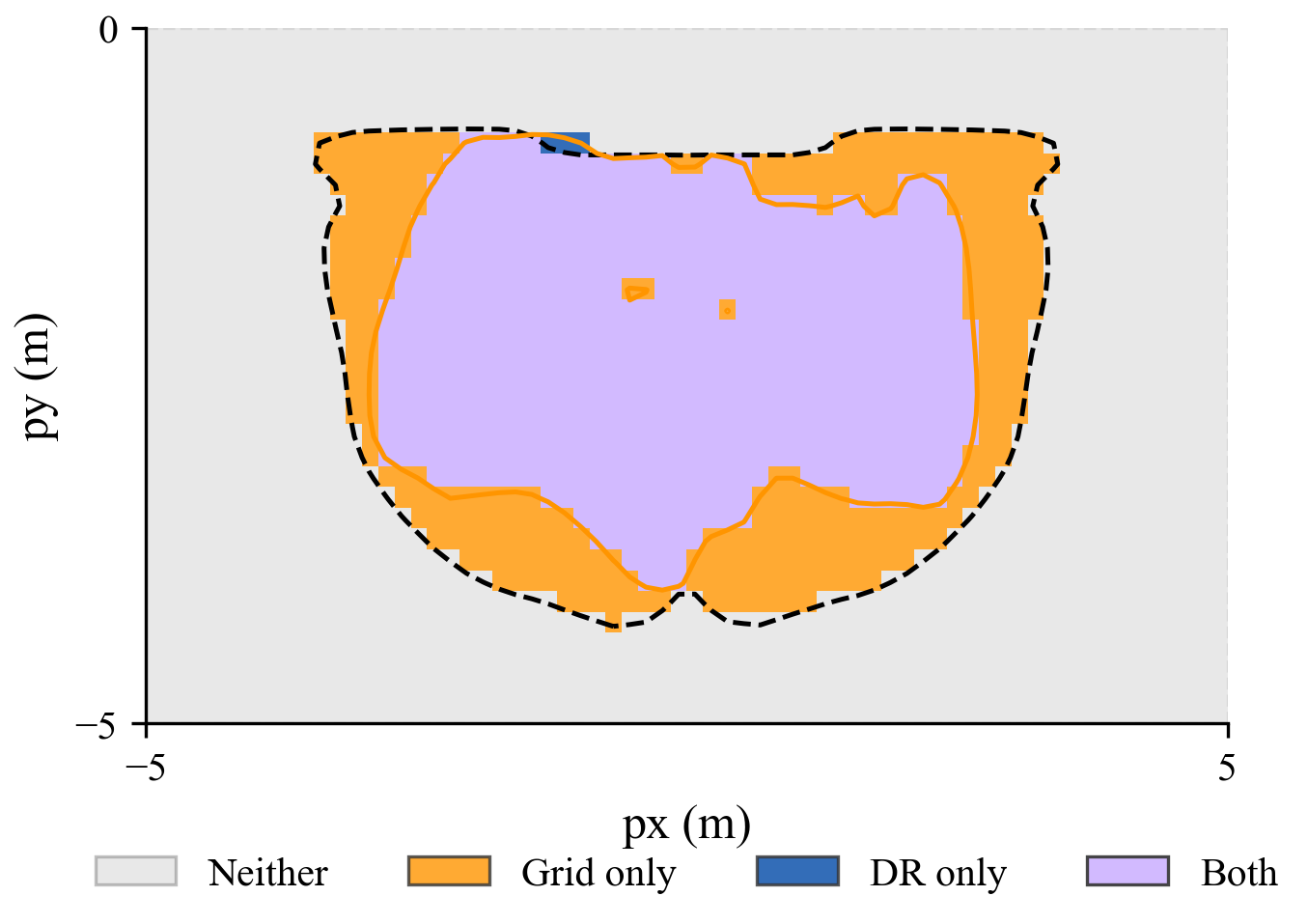}
    \caption{Positional slice BRAT overlap comparison. The learned zero-level set (solid) closely matches the grid-based solution (dashed).}
    \label{fig:brat_overlap}
\end{figure}

\section{Results}

\subsection{Evaluation Overview \& Setup}

The objective of our evaluation is fourfold. First, we evaluate the geometric accuracy of the learned value function against grid-based ground truth. Second, we assess the safety and liveness of the resulting control policy. Third, we quantify the framework's offline and online computational efficiency. Finally, we compare our approach against existing optimal and learning-based controllers. To answer these questions, we evaluate across two case studies: a 6D planar model and the full 13D orbital model.

We evaluate all controllers via closed-loop Monte Carlo simulations. Initial conditions are sampled uniformly from the state space and are filtered to exclude states inside the obstacle or already at the goal. Each simulation runs until docking success, collision, or a timeout of 90s in 6D, and 120s in 13D.

We report: \emph{docking rate}, \emph{collision rate}, \emph{timeout rate}, \emph{mean docking time}, \emph{mean control effort} $\left(\sum \|u\| \Delta t\right)$ over successful trials, \emph{mean time per step} (single control computation), and \emph{mean wall time} (total simulation duration including all steps). All experiments were performed on a workstation equipped with an AMD Ryzen Threadripper PRO 5965WX (24-core) CPU, 128GB RAM, and two NVIDIA RTX 4090 GPUs.
\begin{figure}[!htbp]
      \centering
      \includegraphics[width=1\linewidth]{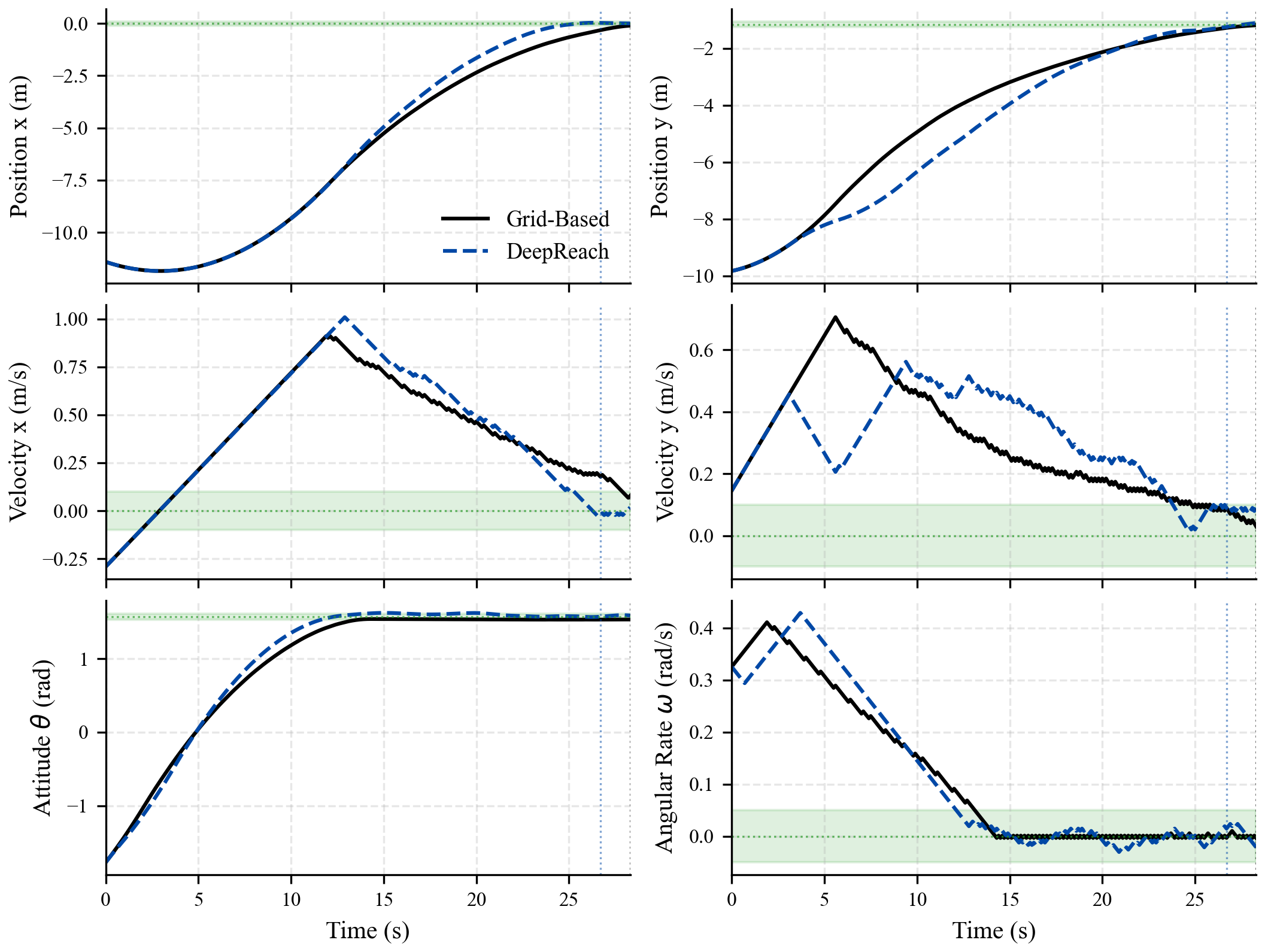}
      \caption{State comparison between grid-based and learned value function controllers for an IC inside the 25\,s grid-based BRAT but outside the 10\,s learned BRAT.}
      \label{fig:state_comparison}
   \end{figure}
\textit{Training Setup.}
The neural value function uses a SIREN MLP~\cite{sitzmann2020siren} with 3 hidden layers of 512 units each. 
The training horizon is $T = 10$\,s for both the 6D and 13D implementations. To ensure robust supervision, the MPC dataset is regenerated periodically. Each regeneration cycle produces $150{,}000$ trajectory rollouts for the 6D model 
and $200{,}000$ rollouts for the 13D model. 
Since the computational cost of data generation scales with the curriculum refinement horizon, we report the timing for the most expensive step: generating the full $10$\,s horizon dataset requires an average of 14 minutes for the 6D model and 46 minutes for the 13D model. Overall, the complete offline training process scales with the complexity of the system dynamics, taking approximately 12 hours for the 6D model and 28 hours for the full 13D model.

\textit{Baselines.}
In addition to the BRAT and Terminal MPC controllers (Sec.~\ref{sec:controllers}), we evaluate four baselines.
\emph{MPC Baseline}: Receding-horizon MPC without a learned value function. This controller uses multi-start gradient shooting with a reach-avoid cost function over a $2$\,s planning horizon.
\emph{Vanilla DeepReach}: Unmodified DeepReach~\cite{bansal2021deepreach} trained with PDE residual minimization alone, without MPC supervision or curriculum refinement.
\emph{RL Reach-Avoid}: The reach-avoid reinforcement learning controller from Hsu et al.~\cite{hsu2021safety}, which learns a reach-avoid value function via model-free RL with safety and liveness guarantees.
\emph{Grid-Based Controller} (6D only): Numerical ground truth via the \texttt{hj\_reachability} library~\cite{hjreachability}. The 6D problem is decomposed into a 4D translational and a 2D rotational subsystem, with combined value function $V_{\mathrm{6D}} = \max(V_{\mathrm{4D}}, V_{\mathrm{2D}})$ solved backward over 30\,s. This decomposition is feasible because the CW translational and single-axis rotational dynamics are decoupled in the planar model.
A BRT safety filter, as defined in Sec \ref{sec:controllers}, was used to supplement all baseline controllers.

\begin{figure*}[!htb]
      \centering
      \includegraphics[width=0.9\linewidth]{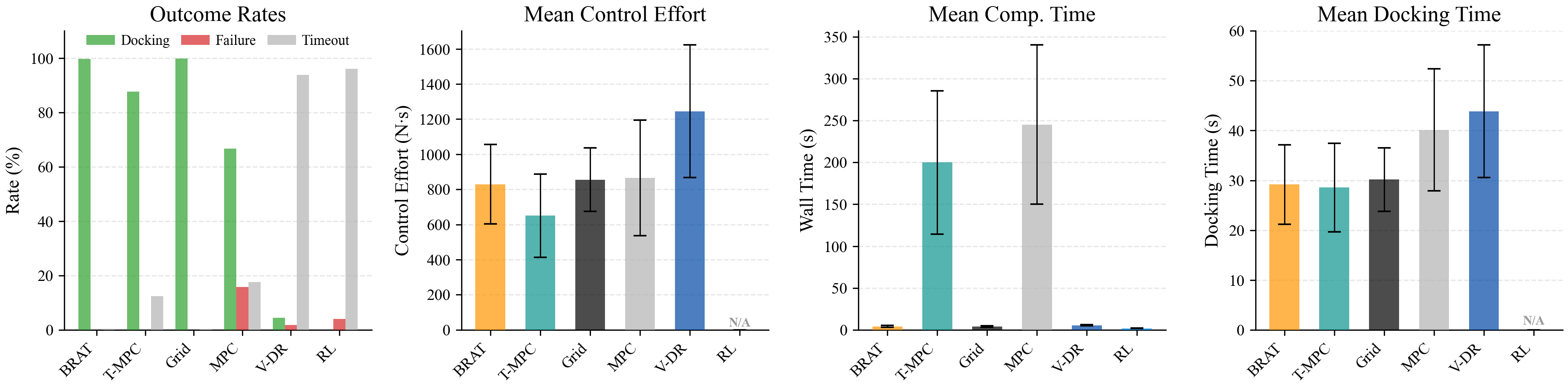}
      \caption{6D controller comparison metrics across all controllers.}
      \label{fig:6d_metrics}
   \end{figure*}

\subsection{Case Study 1: 6D Planar Docking}
The 6D state is $x = [p_x, p_y, v_x, v_y, \theta, \omega]^\top$, with Clohessy--Wiltshire translational dynamics~\cite{clohessy1960} and single-axis rotational kinematics, under planar forces $|u_{x,y}| \leq 20$\,N and torque $|u_\theta| \leq 1.5$\,N$\cdot$m. The target body is a $6 \times 3$\,m rectangle with a docking port extending below it. The \emph{goal set} consists of states where the chaser is (i)~positioned just below the docking port, (ii)~nearly stationary relative to it, and (iii)~aligned in orientation and angular rate for capture (goal attitude $\theta_g = \pi/2$). Formally, this is encoded via the reach function $g(x) = \max(d_{\mathrm{pos}}, d_{\mathrm{vel}}, d_{\mathrm{att}}, d_{\mathrm{rate}})$, where each $d_i \leq 0$ indicates satisfaction of the corresponding tolerance ($\varepsilon_p = 0.1$\,m, $\varepsilon_v = 0.1$\,m/s, $\varepsilon_\theta = 0.04$\,rad, $\varepsilon_\omega = 0.05$\,rad/s). The \emph{failure set} is the target body and port, each inflated by the chaser's bounding-circle radius $r_c \approx 0.707$\,m. Since the CW dynamics are control-affine with constant $B$, the bang-bang control~\eqref{eq:bangbang} is read directly from the value function gradient.

We validate the learned 6D value function against the decomposed grid-based solution and compare closed-loop performance through four complementary analyses.

\textit{1) Volumetric Accuracy.}
Before evaluating closed-loop control performance, we assess the geometric fidelity of the learned value function. Visually, Fig.~\ref{fig:brat_overlap} compares a positional slice of the learned and grid-based BRATs, showing that the learned zero sub-level set closely approximates the grid-based boundary. To quantify this, we treat the zero sub-level set as a binary classifier: a state is labeled \emph{safe} (positive) if it lies inside the BRAT ($V \!\leq\! 0$) and \emph{unsafe} (negative) otherwise. We draw $5 \!\times\! 10^5$ states uniformly from a focused 6D evaluation region centered on the BRAT. This focused region ensures adequate sampling density near the reachable boundary. Table \ref{tab:volumetric} compares each approximation against the grid-based value $V_{\mathrm{grid}}$ at $t = 0$~s and report the True Positive Rate (TPR) and False Positive Rate (FPR). A high TPR indicates the learned BRAT is not overly conservative, recovering most of the reachable set. Conversely, a high FPR reveals over-optimism: the method incorrectly claims unreachable states as safe, which can weaken safety guarantees.

\begin{table}[h]
\caption{6D Volumetric Accuracy vs.\ Grid ($t=0$\,s, $N=5{\times}10^5$)}
\label{tab:volumetric}
\centering
\begin{tabular}{lcc}
\toprule
Method & TPR (\%) $\uparrow$ & FPR (\%) $\downarrow$ \\
\midrule
Vanilla DeepReach        & 10.84 & \textbf{0.01} \\
RL BRAT                  & \textbf{95.58} & 58.52 \\
Proposed DeepReach (ours) & 81.12 & 0.15 \\
\bottomrule
\end{tabular}
\end{table}

\begin{remark}
The grid-based baseline suffers from numerical dissipation and discretization errors around sharp geometries like the docking port. Consequently, some reported FPs and FNs may actually reflect the continuous network resolving gradients that the coarse grid blurs out. We therefore rely on closed-loop Monte Carlo rollouts to definitively evaluate controller safety and liveness.
\end{remark}
\begin{figure}[H]
  \centering
  \includegraphics[width=0.9\linewidth]{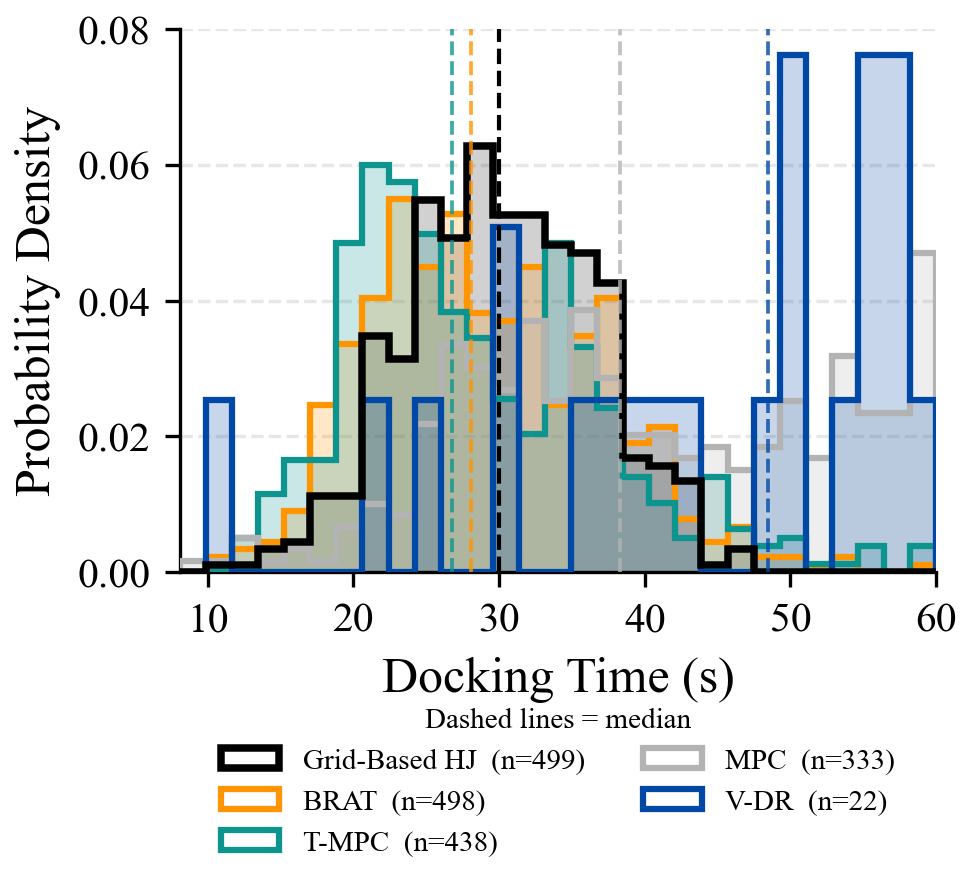}
  \caption{6D docking time histogram. The learned BRAT controller achieves faster docking times than the grid-based solution due to discretization errors in the grid-based value function.}
  \label{fig:6d_docking_hist}
\end{figure}

\begin{figure*}[!htbp]
    \centering
    \includegraphics[width=0.9\linewidth]{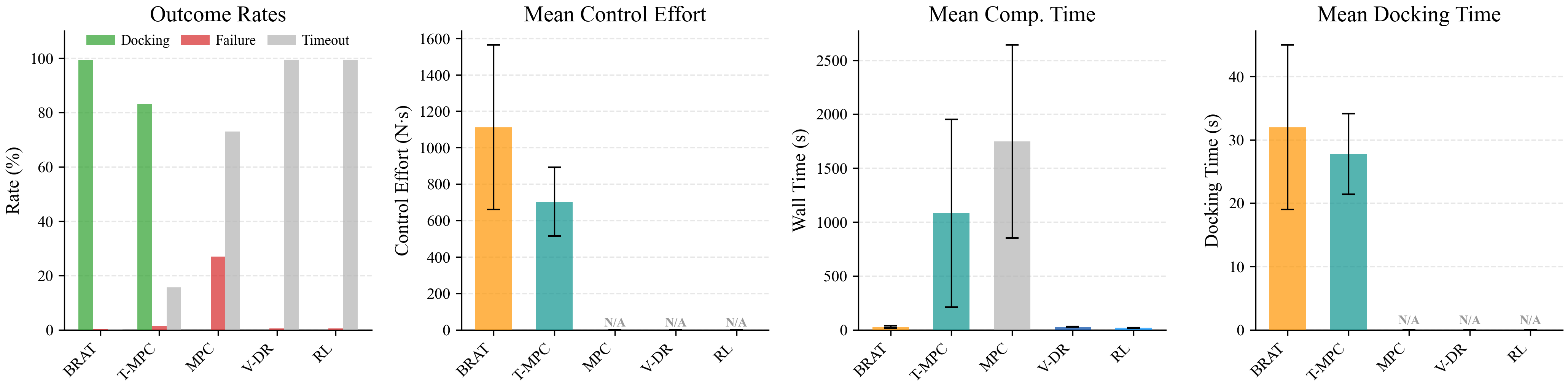}
    \caption{13D controller comparison metrics across all applicable controllers.}
    \label{fig:13d_metrics}
\end{figure*}

\textit{2) Trajectory Optimality.}
Fig.~\ref{fig:state_comparison} compares closed-loop trajectories from the learned BRAT controller and the grid-based controller, starting from an initial condition inside the 25\,s grid-based BRAT but outside the 10\,s learned BRAT. Despite operating with a shorter effective horizon, the learned controller closely tracks the grid-based trajectory across most state dimensions, demonstrating that the learned value function gradient produces near-optimal control actions even outside the training horizon.


\textit{3) Rollout Analysis.}
Table~\ref{tab:brat_rollouts} (6D column) reports $10{,}000$-sample rollout results evaluating the controller from uniformly sampled initial conditions (ICs). By drawing ICs uniformly from the full evaluation state space with the least-restrictive safety filter enabled, we demonstrate that the proposed controller effectively docks from arbitrary states—including those outside the trained BRAT. Crucially, this approach achieves high success rates without requiring a long training horizon, setting the stage for outperforming both numerical and learning-based baselines.

\begin{table}[!htbp]
\caption{BRAT Rollout Results ($N = 10{,}000$ uniformly sampled ICs)}
\label{tab:brat_rollouts}
\centering
\begin{tabular}{lcccccc}
\toprule
       & Docking & Collision & Timeout & Time/step & Effort \\
       & (\%)    & (\%)      & (\%)    & (ms)       & (N$\cdot$s) \\
\midrule
6D  & 99.94 & 0.01 & 0.05 & 10.74  & 780.97  \\
13D & 98.88 & 0.58 & 0.54 & 16.00 & 1115.41 \\
\bottomrule
\end{tabular}
\end{table}

\textit{4) Controller Comparison.}
Table~\ref{tab:6d_comparison} compares all controllers over $500$ shared initial conditions, and Fig.~\ref{fig:6d_metrics} summarizes the results. The proposed BRAT controller achieves a $99.6\%$ success rate, matching the performance of the grid-based ground truth ($99.8\%$) while maintaining a strict $0.0\%$ collision rate. In contrast, the receding-horizon MPC baseline struggles with the non-convex reach-avoid landscape, yielding a $15.8\%$ collision rate and only $66.6\%$ success. The Vanilla DeepReach and RL baselines fail almost entirely (predominantly via timeout), underscoring the necessity of the proposed MPC supervision and curriculum refinement. While the Terminal MPC (T-MPC) achieves the lowest overall control effort ($650.47$\,N$\cdot$s), it sacrifices online computational efficiency. The pure BRAT controller operates with minimal computational overhead---requiring only a single forward pass through the SIREN network and one backward pass to compute the gradient at each control step---achieving a real-time execution speed of $10.74$\,ms per step. In contrast, T-MPC and standard MPC require orders of magnitude more time per step due to online optimization overhead.

\begin{table}[!htbp]
\caption{6D Controller Comparison ($N = 500$ rollouts)}
\label{tab:6d_comparison}
\centering
\small
\setlength{\tabcolsep}{3pt}
\begin{tabular}{lcccccc}
\toprule
Metric & BRAT & T-MPC & Grid & MPC & V-DR & RL \\
\midrule
Success (\%)       & 99.6   & 87.6   & \textbf{99.8}   & 66.6   & 4.4    & 0.0 \\
Collision (\%)     & \textbf{0.0}    & \textbf{0.0}    & \textbf{0.0}    & 15.8   & 1.8    & 4.0 \\
Timeout (\%)       & 0.4    & 12.4   & \textbf{0.2}    & 17.6   & 93.8   & 96.0 \\
Effort (N$\cdot$s) & 829.10 & \textbf{650.47} & 854.89 & 866.32 & 1244.97 & N/A \\
Time/step (ms) & 10.74 & 2246.85 & 13.91 & 5880.90 & 5.86 & \textbf{0.35} \\
Dock time (s)      & 29.17  & \textbf{28.57}  & 30.18  & 38.30  & 48.45  & N/A \\
\bottomrule
\end{tabular}
\end{table}

\textit{5) Docking Time Optimality.}
Fig.~\ref{fig:6d_docking_hist} compares docking time distributions across all controllers against the grid-based BRAT. Interestingly, the learned BRAT controller achieves marginally faster mean docking times ($29.17$\,s) compared to the numerical ground truth ($30.18$\,s). This superior optimality can be attributed to the continuous nature of the neural value function, which mitigates the artificial dissipation and discretization errors inherent in the grid-based solution, particularly when resolving the tight tolerances of the docking port goal set.

\subsection{Case Study 2: 13D Orbital Docking}
The full model adds a third spatial axis, unit-quaternion attitude, and three-axis angular velocity, giving $x \in \mathbb{R}^{13}$. Translational dynamics follow the 3D Hill--Clohessy--Wiltshire equations~\cite{clohessy1960, alfriend2010} with body-to-LVLH force rotation; rotational dynamics follow Euler's equations with quaternion kinematics. Control comprises body-frame forces $|F_i| \leq 20$\,N and torques $|\tau_i| \leq 1.5$\,N$\cdot$m per axis. The goal set follows IDSS docking tolerances~\cite{nasaidss}: lateral position 
$\leq 0.10$\,m, lateral velocity $\leq 0.02$\,m/s, axial approach velocity $\in [0.03, 0.10]$\,m/s, attitude error $\leq 8.7^\circ$, and angular rates $0.15$--$0.4^\circ$/s depending on axis. The failure set uses orientation-dependent inflation along the docking 
axis via the chaser cube's support function, allowing the goal band to sit closer to the port when the chaser is properly aligned. 
Thrust is applied in the body frame while the value function gradient is expressed in LVLH, so the effective force coefficient is $R(\mathbf{q})\,\nabla_v V / m_c$ and the torque coefficient is $I^{-\top}\nabla_\omega V$, with the bang-bang sign~\eqref{eq:bangbang} taken on these transformed quantities.

\textit{1) Rollout Analysis.}
Table~\ref{tab:brat_rollouts} (13D row) reports $10{,}000$-sample rollout results under the same IC regime as in 6D. The grid-based controller is excluded because the 13D state space is computationally intractable---a $51$-point grid per dimension would require $51^{13} \approx 10^{22}$ grid points. The proposed BRAT framework scales successfully to the full translational-rotational coupled system, achieving a $98.88\%$ docking success rate and maintaining a sub-$1\%$ collision rate across arbitrary initial conditions. This demonstrates that the learned value function effectively generalizes the reach-avoid guarantees to a high-dimensional state space without requiring exhaustive and computationally intractable grid discretization.

\begin{figure}[!htbp]
      \centering
      \includegraphics[width=0.95\linewidth, trim=0 0 0 25cm, clip]{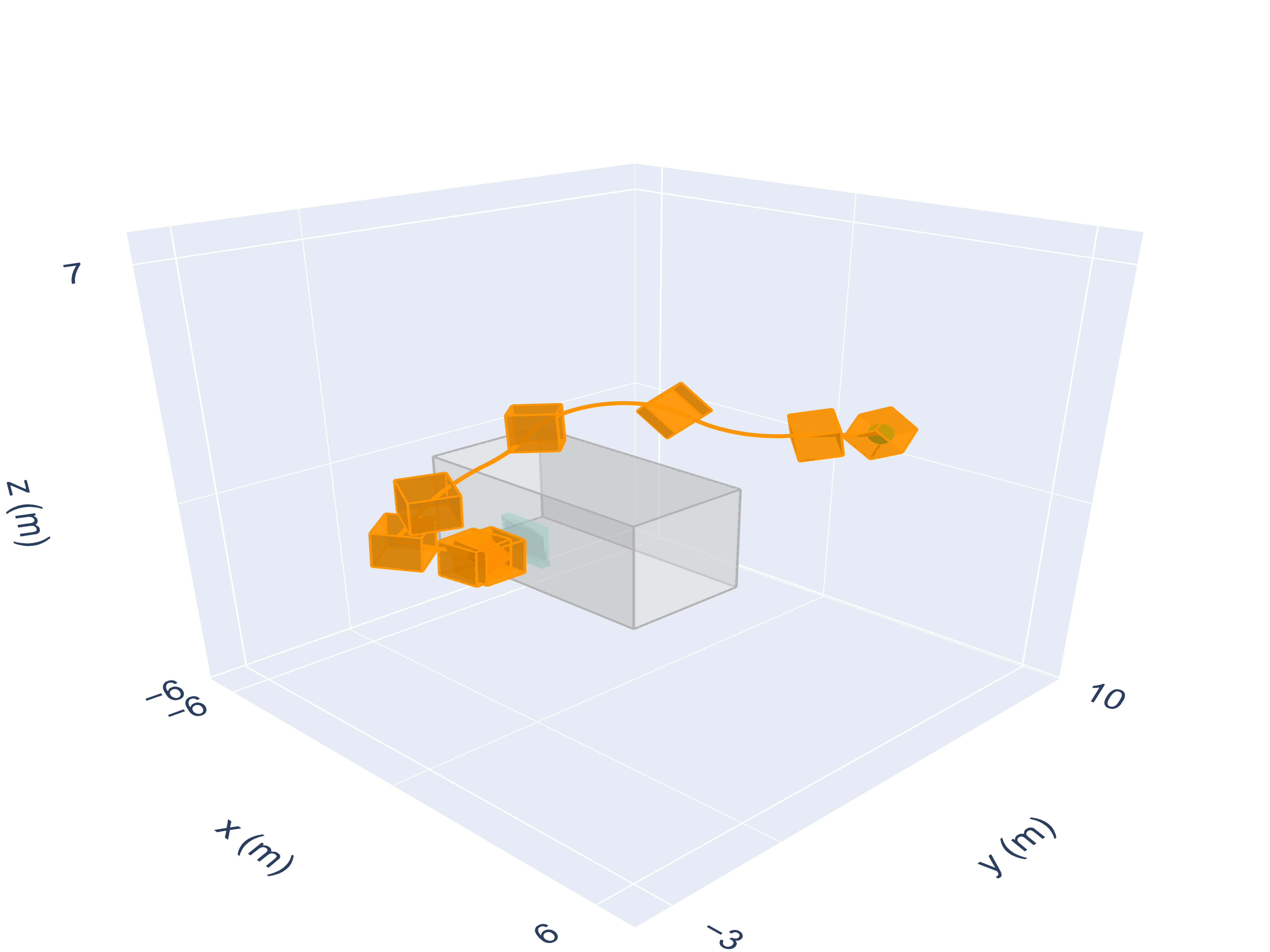}
      \caption{Example 13D trajectory starting outside the learned BRAT, demonstrating successful docking with the least-restrictive safety filter.}
      \label{fig:13d_trajectory}
   \end{figure}

\textit{2) Controller Comparison.}
Table~\ref{tab:13d_comparison} compares all applicable controllers over 500 shared initial conditions, and Fig.~\ref{fig:13d_metrics} summarizes the results. Fig.~\ref{fig:13d_trajectory} shows an example 13D trajectory starting outside the learned BRAT, demonstrating successful docking with the least-restrictive safety filter. The results highlight a stark performance gap in high dimensions. While the proposed BRAT controller achieves a near-perfect success rate (99.2\%), the baselines struggle significantly to balance safety, liveness, and computation. The Terminal MPC (T-MPC) controller manages an 83.0\% success rate but requires over 12 seconds (12690.58 ms) per control step, rendering it impractical for real-time onboard execution. Standard MPC completely fails to solve the highly constrained 13D geometry, colliding in 27.0\% of trials and timing out in the rest. Similarly, the Vanilla DeepReach and RL baselines fail to synthesize meaningful progress toward the goal, resulting in 99.4\% timeout rates. By contrast, the BRAT controller maintains strict real-time execution, requiring only 16.00 ms per control step, providing a computational efficiency and reliability that optimization-based and baseline learning approaches cannot match.

\begin{table}[!htbp]
\caption{13D Controller Comparison ($N = 500$ rollouts)}
\label{tab:13d_comparison}
\centering
\begin{tabular}{lccccc}
\toprule
Metric & BRAT & T-MPC & MPC & V-DR & RL \\
\midrule
Success (\%)       & \textbf{99.2} & 83.0 & 0.0 &  0.0 & 0.0 \\
Collision (\%)     & \textbf{0.4} & 1.4 & 27.0 & 0.6 & 0.6 \\
Timeout (\%)       & \textbf{0.4} & 15.6 & 73.0 & 99.4 & 99.4 \\
Effort (N$\cdot$s) & 1112.06 & \textbf{703.51} & N/A & N/A & N/A \\
Time/step (ms)     & 16.00 & 12690.58  & 19569.50 & 5.93 & \textbf{0.39}  \\
Dock time (s)      & 31.96 & \textbf{27.74} & N/A & N/A & N/A \\
\bottomrule
\end{tabular}
\end{table}

\balance
\section{Conclusion}

We presented a learning-based BRAT framework for autonomous spacecraft docking in 13 dimensions, combining neural value function approximation with MPC-generated curriculum labels. The framework follows an offline--online deployment strategy: the expensive training and MPC data generation are performed once, producing a compact SIREN network that is queried in real time with only a single forward and backward pass per control step, eliminating the need for onboard optimization. In 6D, the learned BRAT controller closely matches the grid-based ground truth while achieving faster docking times, attributable to discretization errors in the grid-based value function. In 13D the BRAT controller achieves a 98.88\% docking rate with a 0.58\% collision rate across 10{,}000 uniformly sampled initial conditions, outperforming MPC, vanilla DeepReach, and RL baselines at substantially lower online computational cost.

\subsection{Limitations and Future Work.}
The primary limitation is that the formulation is disturbance-free and the target satellite is stationary (non-tumbling) in the LVLH frame. The grid-based framework supports bounded disturbances via the differential game formulation, but this extension remains to be incorporated into the neural approach. Recent work on adversarial DeepReach~\cite{teoh2025} provides a path toward this by extending MPC-guided training to two-player zero-sum games, and integrating such robustness into the BRAT pipeline is a natural next step. 

To address the convergence limitation, we plan to investigate neural network verification techniques adapted to high-dimensional value functions. Robustness to disturbances can be incorporated by extending the Hamiltonian to include a maximizing adversary, following the standard differential game formulation~\cite{mitchell2005}, with the neural training pipeline adapted accordingly. Extension to tumbling targets requires a time-varying goal and failure set, addressable within the HJ framework by coupling target attitude dynamics into the state~\cite{zagaris2018}.




\bibliographystyle{ieeetr}
\bibliography{refs}

\end{document}